\documentclass{ecai} 
\usepackage{latexsym}
\usepackage{amssymb}
\usepackage{amsmath}
\usepackage{amsthm}
\usepackage{booktabs}
\usepackage{enumitem}
\usepackage{graphicx}
\usepackage{color}
\usepackage{soul}
\usepackage{url}
\usepackage{xcolor}
\usepackage[hidelinks]{hyperref}
\usepackage[utf8]{inputenc}
\usepackage[small]{caption}
\usepackage{graphicx}
\usepackage{natbib}
\usepackage{algorithm}
\usepackage{algorithmic}
\newtheorem{definition}{Definition}
\newtheorem{prop}{Proposition}

\newcommand{\BibTeX}{B\kern-.05em{\sc i\kern-.025em b}\kern-.08em\TeX}

\newcommand{\fh}{\hat{f}}

\newcommand{\hidden}[1]{}

\begin{document}

\begin{frontmatter}
 
%%% Use this command to specify your submission number.
%%% In doubleblind mode, it will be printed on the first page.

%\paperid{1904}

%%% Use this command to specify the title of your paper.

\title{On explaining with attention matrices}

\author[A]{\fnms{Omar}~\snm{Naim}\thanks{Corresponding Author. Email: omar.naim.docs@gmail.com}}
\author[B]{\fnms{Nicholas}~\snm{Asher}\thanks{Corresponding Author. Email: nicholas.asher@irit.fr}}

\address[A]{Université Toulouse 3, IRIT France}
\address[B]{CNRS, IRIT France}

%%% Use this environment to include an abstract of your paper.
\begin{abstract}
This paper explores the much discussed, possible explanatory link between attention weights (AW) in transformer models and predicted output. Contrary to intuition and early research on attention, more recent prior research has provided formal arguments and empirical evidence that AW are not explanatorily relevant.  We show that the formal arguments are incorrect.  We introduce and effectively compute {\em efficient attention}, which isolates the effective components of attention matrices in tasks and models in which AW play an explanatory role.  We show that efficient attention has a causal role (provides minimally necessary and sufficient conditions) for predicting model output in NLP tasks requiring contextual information, and we show, contrary to \cite{brunner:etal:2020}, that efficient attention matrices are probability distributions and are effectively calculable. Thus, they should play an important part in the explanation of attention based model behavior.  %We start from \cite{brunner:etal:2020}'s demonstration that AW are not identifiable in some realistic cases. 
%We argue that identifiability of AW is important for their role in faithful and plausible explanations.  
%  Our new concept of efficient attention solves an important problem with \cite{brunner:etal:2020}; their solution to ensure AW identifiability is not necessarily a probability distribution, threatening its explanatory value.  We prove that efficient attention ensures identifiability of AW and  remains a probability distribution while being effectively calculable.  %{\color{magenta}\cite{jain:wallace:2019, wiegreffe:pinter:2019} look at AW and explainability, \cite{brunner:etal:2020} look at AW and identifiability; In our paper, we investigate the relationship between AW, explainability and identifiability.}
We offer empirical experiments in support of our method illustrating various properties of efficient attention with various metrics on four datasets.
%The code and datasets used for our experiments can be found at \url{https://github.com/omyokun/On-explaining-with-attention-matrices}
\end{abstract}

\end{frontmatter}

\section{Introduction}

Transformer and other attention based models optimize attention weights (AW), the distribution of weights in transformer models' implicit representations of the hidden states of tokens, as inputs to multilayer perceptrons (MLPs) in order to solve a variety of tasks.  As such it stands to reason that AW should have an explanatory role \cite{cho:etal:2014,ghaeini:etal:2018,vig:belinkov:2019,clark:etal:2019}. %\cite{vig:belinkov:2019,clark:etal:2019} used AW to explain a model's predictions, by considering the inputs with the greatest weights as responsible for the output. 
However, many have criticized using AW to explain model predictions \cite{serrano:smith:2019,jain:wallace:2019,wiegreffe:pinter:2019,galassi:etal:2020,brunner:etal:2020,bibal:etal:2022}.   
\cite{brunner:etal:2020} formally proved that under certain assumptions an infinite number of AW could yield the same prediction; i.e., AW are {\em unidentifiable} from output--the relation from AW to predictions is not 1-1.  Empirical testing by \cite{jain:wallace:2019,wiegreffe:pinter:2019,vashishth:etal:2019,de-santana:colombini:2022} has confirmed non-identifiability of AW in at least some tasks and models, which \cite{arous:etal:2021,chrysostomou:aletras:2021} among others cite as a drawback for explanability.%On the other hand, \cite{wiegreffe:pinter:2019} suggest that the link between identifiability and explainability is not needed.  %In addition,  have shown that AW do seem identifiable in some NLP tasks \cite{},    

This paper makes three main contributions.  First, we bring together in a novel way, AW, identifiability and explainability in transformer models, linking identifiability of AW with counterfactual and causal explanations and explanatory faithfulness
\cite{rudin:2019,jacovi:goldberg:2020,yin:neubig:2022}.  

Second, we show that  \cite{brunner:etal:2020}'s approach fails to  %***We argue that while a lack of identifiability does not preclude faithful explanations {\em per se}, identifiability strengthens explanations and makes them more compelling.***  Using the tool of formal interpretability, we show why such strengthening is expected. 
make AW viable explanatory tools. While they identify a projection into a representation space that looks promising they are unable to show that the weights in that projection have the properties of a probability distribution.  Moreover, they do not see a way to calculate the result of the projection effectively into a space with the requisite properties.  This vitiates twice over the explanatory power of their solution.  We introduce a new projection whose result we call {\em, efficient attention}.  Efficient attention restores identifiability; and we show that its weights define a probability distribution. Finally, we show how to calculate this projection effectively.

Third, we present a series of experiments\footnote{https://github.com/omyokun/On-explaining-with-attention-matrices/} on various datasets using the experimental set up of \cite{wiegreffe:pinter:2019,jain:wallace:2019}. The point of these experiments is to show empirically the effects of efficient AW.  We show first that AW matrix makes the same predictions as its efficient AW projection.  Second, we take AW and their adversarial AW from \cite{wiegreffe:pinter:2019} with the same predictions, and we show empirically that they have the same efficient AW projection within the limits of integer precision. Finally, we show that intervening on model and replacing one efficient AW with another shifts the predictions, confirming empirically the identifiability of efficient AW and their potential role in counterfactual, faithful explanations, contrary to \cite{jain:wallace:2019,wiegreffe:pinter:2019,chrysostomou:aletras:2021}. %and \href{https://anonymous.4open.science/r/On-Explaining-Transformers-with-Attention}{here}.
% This projection isolates the components of AW that are causally relevant for predictions and restores AW as causally relevant, explanatory devices, .  

 %Brunner et al provide here the base version of an effective attention, and leave the investigation of $Ker([T,1]^{T})^{T}$ for future research. \\

Section \ref{sec:background} starts with an analysis of explainability and identifiability.  Section \ref{sec:math} provides a detailed look at the theory behind the claims of  \cite{brunner:etal:2020} about non-identifiability of AW and their solution.  We introduce {\em efficient attention}, prove its identifiability and that it is a probability distribution. We also show how to calculate it.  Section \ref{sec:empirical} empirically validates the identifiability of efficient attention contrary to the claims of \cite{wiegreffe:pinter:2019} using their experimental set up and datasets.  % in particular we take two matrices from \cite{wiegreffe:pinter:2019} with the same prediction and show they have the same efficient attention core. 
%We then conclude. 

\section{Explainability and Interpretability for Transformers} \label{sec:background}

 \paragraph{Attention and transformers}  Generative models or transformers with decoders predict the next token $y_{i+1}$ using the conditional probability $P(y_{i+1} | y_0 , \dots , y_{i}, C_{i+1})$ where $y_0$ is the seed vector, $\{y_1 , \dots , y_i\}$ are prior predicted tokens and $C_{i+1}$ is a latent representation of $y_{i}$ together with other information given to the model. Encoder transformer and other attention based models also use these latent representations of input tokens.  % state for each previously predicted token and tokens of $C$.  %$p(y_{i+1}/y_1, \dots, y_{i}, c_i)$. 
 %Hence, the output $y$ = \{$y_1 , \dots, y_{T}$\} is directly linked to the calculation of this probability, which depends on the contextualization vectors.  
%In the case of transformers, $Attention(Q,K,V)H = A.V.H = AT =  [c_1,..,c_N]$.
Attention matrices or their AW determine these representations $C_i$ from entries that are typically already vector encodings.  %, which in turn determine the final output of the model. 
More specifically, $C_i$ is a weighted sum of $C_i^{h}$ over $h$ attention heads, where each $C_i^{h}$ is: %that  with 
\begin{equation}
C^{h}_i = \Sigma^K_{k=1}\alpha^{h}_{i,k}(W^{h}_Ve_k)  \label{attn}
    \end{equation}
$\alpha_{i,k}^{h}$ is the attention weight providing the import of token representation $e_k$ to $e_i$ and $W^{h}_V$ is the value weight matrix of the head $h$. 

%AW are part of a transformer or other attention based model model.  
A typical transformer has multiple AW that link with a residual stream from the previous layer to a multi-layer perceptron (MLP).  %\cite{jain:wallace:2019,wiegreffe:pinter:2019} ran experiments on attention matrices that were not embedded in the full transformer architecture.  
Since early work on AW (\cite{jain:wallace:2019,wiegreffe:pinter:2019}  did not even use full transformer models), we have learned a lot more about how transformer models exploit context input tokens and AW.  The Transformer architecture has layers with each having different blocks in the following order: attention, residual learning \cite{he2016deep}, layer normalization \cite{ba2016layer} and MLP (Multi Layer Perceptron). \cite{kobayashi:etal:2021,geva2021transformer,ferrando:etal:2022, ferrando:etal:2023} show how to decompose attention blocks in a way that allows us to interpret token/token interactions in each block.  AW rearranges the input embeddings in a transformer model to optimize the internal representation input to the MLP that must execute some task.  The role of residual stream is to incorporate elements from previous layers to avoid overfitting at each layer; but this feeds into the next block's attention layer.  The role of the layer normalization block is to significantly reduce training time. Finally the MLP block learn the function we want from the transformer.

All of this work points to the important role of attention in transformer model predictions.  If attention did not play an explanatory role, by giving a better contextualization to facilitate the learning for the MLP, the Transformer performance would reduce to that of the MLP component, but this is manifestly not the case. \cite{perez:etal:2021}shows the importance of the attention layer in their theoretical analysis of the computational power of transformer architectures. Thus, the optimized representation of input data by the attention layer should have an important causal effect on the MLP's computations and the eventual output, in any task where the first input to the transformer model is not sufficient for an MLP to solve the task. We know from formal results about classifiers that Heavyside MLPs \citep{duff:etal:1966} benefit greatly from a representation space that minimizes the number of ``jumps'' the classification function must make in order to classify some data.\footnote{For a discussion, see https://nitishpuri.github.io/posts/books/a-visual-proof-that-neural-networks-can-compute-any-function/.}  Moreover, the effects of AW at different layers of a transformer model may be different, as in circuit analysis \cite{rauker:etal:2023}, mechanistic interpretability \cite{olah:etal:2020,meng:etal:2022,geiger:etal:2021}. %However, prior research \cite{jain:wallace:2019,wiegreffe:pinter:2019,brunner:etal:2020} has provided both formal and empirical arguments that AW cannot play this role.  

We grant that not all tasks may need the contextual information of equation \ref{attn}, at least not in all layers or in the same way.  For instance, simple classification tasks like gender identification do not seem to require the contextual encodings of attention heads, at least not through all layers \cite{jourdan:etal:2023a}.  \hidden{However, even in cases where the model doesn't exploit the contextualization vectors, attention layers in transformer architectures play an important role in projecting the tokens into a fixed dimension required for MLP input.}
Nevertheless, it stands to reason that AW should play an important explanatory role.  We will show, contrary to \cite{jain:wallace:2019,wiegreffe:pinter:2019,brunner:etal:2020}, that AWs are causally determinative of the output and so should in principle have explanatory value.

   %as a projection of the tokens into a fixed dimention needed for the MLP input.%, in which case AW will not play an explanatory role \cite{wiegreffe:pinter:2019}. 

\paragraph{Explainability} With this in mind, we turn to explainability with AW.  Discussions of explainability for transformers often start from different interpretations of explainability and interpretability.  \cite{jacovi:goldberg:2020} use them equivalently, but \cite{rudin:2019} distinguishes them using the concepts of faithfulness and plausibility.  Plausibility has to do with how acceptable the purported explanation is to humans \cite{jacovi:goldberg:2020}.   Faithfulness involves a causal connection, what is causally necessary and sufficient for the model to produce its prediction given a certain input; a faithful explanation is one that captures the causal relations in the explanations it generates of a model's predictions.  That means a faithful explanation should ideally identify minimally sufficient and necessary  conditions for the prediction.   

A faithful explanation of the behavior $\fh$ thus has to support counterfactual interventions of the form: $A$ {\em and} $B$ and {\em had $A$ not been the case, $B$ would not have been the case either} \cite{jacovi:goldberg:2020}. %We wll model these causal explanations with counterfactual explanations . 
Using \cite{Lewis:1981}'s definition of causation and given $\fh:X \rightarrow Y$, we say that $A$ faithfully explains why $f(x) = \pi$ for $x \in X$ iff we can establish (i) $x$ has $A$ ($A(x)$), and (ii) if $A(x)$ hadn't been the case, $\fh(x)\neq \pi$.  

To establish such counterfactuals we typically need a background theory $T$ that could be formal  (as in \cite{tarski:etal:1953}), but it could also be based on observation.  $T$ requires a range of {\em cases} $\Phi$, different set ups or variations on inputs that support counterfactual interventions, in which we remove our putative explanatory factors $\phi$ from $x$ but keep the rest of our set up $\fh$ and the other properties of $x$ the same or as similar as possible.  Counterfactuals require a distance metric over $\Phi$ \cite{lewis:1973} to make precise the notion of similarity.  The appropriate metric depends on the nature of the cases and the task.  There are many candidates for counterfactual theories of a transformer $\fh$ \cite{asher:etal:2022}. One can set $\Phi$ to the feature space of the inputs to the model.  Alternatively we can set $\Phi$ to a set of possible AW in $\fh$ or $W = $ a set of possible parameter settings for the final layer of $\fh$ as suggested in \cite{wiegreffe:pinter:2019,fel:etal:2023},  or any among a large number of other possibilities of parameter values for intermediate states in $\fh$.  Not all of these candidates may provide faithful explanations, however, a choice of cases may not turn out to support any nontrivial counterfactuals.

To have faithful explanations and non trivial counterfactuals, we need identifiability of the explanatory factors from the predictions.  The non-identifiability of AW precludes counterfactual forms of explanation of the form {\em had the AW been different, the prediction would have differed}, and thus threatens the faithfulness of any explanation based on AW.  However, there are a couple of caveats.  In general, in light of finite integer precision and the approximation of exact real values and computations in transformers, we should not expect small variations in AW numerical values to practically affect prediction.  %Since exact numerical values are not possible, we have to expect that small variations in the numerical values of AW might not shift predictions.
We are interested in the possibility of faithful explanations using sets ${\cal A}$ of AW where  for $x, y \in {\cal A}, \| x - y\| < \lambda $ for $\lambda$ determined by the task, approximation and integer precision, the predictions are more or less the same and where for distinct ${\cal A, A}'$ and for $x \in {\cal A}, y \in {\cal A}'$ with $\| x - y\| > \lambda $,  we should get distinct predictions.  \cite{jain:wallace:2019,wiegreffe:pinter:2019,vashishth:etal:2019} provide empirical evidence that at least in some NLP tasks, what appear to be numerically significant shifts in AW do not affect the predictions over a large number of instances in an NLP task.

If the variations in AW values are large and are not explicable in terms of approximations, then the explanatory plausibility carried by such a large set of AW is much less.  Such sets, which are a mark of causal overdetermination \cite{loeb:1974}, leave humans wanting more: what is the common element that makes all of these cases causally relevant?    In some  NLP tasks, for instance, text classification, this overdetermination might be innocuous or even indicative that the task does not distinguish between a broader class of AW numerical values; a single text input might provide many sufficient clues for a classification; a single movie review might be classified as favorable because of several passages, a high attention value on any one of which would suffice for a favorable classification.  But in general, given a particular input, identifiable AW distributions (up to a certain arithmetic precision) for a particular prediction would carry a higher explanatory value in terms of plausibility than sets of heterogeneous AW values. There is a point then to searching for identifiability of AW within the limits of integer precision. Identifiable attention matrices can also furnish the basis for more sophisticated strategies for plausible explanations with attention as in \cite{heo:etal:2020}.

There is another dimension of plausibility to consider.  A plausible explanation using AW needs to link AW to input tokens.  Recalling what attention layers do, they provide ``mixtures'' of input token effects on each token.  These mixtures are difficult if not impossible to interpret when the values are negative, and they are most easily interpretable when they are probability distributions as motivated in a different context by \cite{lee:seung:1999}.  That is, attention weights as probability distributions provide {\em a priori} for humans more plauible explanations.

\section{Mathematics for Attention and Identifiability} \label{sec:math}

 \cite{brunner:etal:2020} showed a sufficient condition for non identifiability: for a given input sequence longer than the attention head dimension, a transformer model can generate an infinite set of attention weight distributions producing the same output prediction. This is not just an abstract possibility; in cases with relatively long inputs, non identifiability is real. \cite{brunner:etal:2020} also proposed a means for removing the weight components that do not influence the model’s prediction to make AW identifiable.  %, as it makes the function from those values to predictions 1-1.  
 But their approach doesn't guarantee that AW are probability distributions. 

To analyze and unpack this problem, we first review and give a detailed proof of \cite{brunner:etal:2020}'s necessary and sufficient condition for identifiability of AW. We then describe their solution of effective attention, which provides a projection of AW values that renders the function from AW values to predictions injective but at the price of making those values no longer a probability distribution.  We then define a new and computable projection of AW into what we call {\em efficient AW} that solves the mathematical problem with the effective attention \cite{brunner:etal:2020} while also restoring identifiability of attention weight distributions.% by eliminating the weight components that have no impact on the model's predictions.

\subsection{Identifiability on AW}
 %Identifiability of AW statistically means that a model $\fh$ is injective of its parameters AW--i.e. the weight distribution is unique for a given output.   %Therefore, if AW are not identifiable, attention-based explanations are no longer warranted since, if it normally explains the output, then any variation in the distribution of attention must vary the output, which is not the case for non-identifiability. \\

As background to the work of \cite{brunner:etal:2020}, multi-headed attention output combine h independent attention heads with reduced head dimension $d_v = d/h$ in parallel fashion through a linear layer. The output is achieved by summing outputs over each single head and multiplying them by the matrix $H \in \mathbb{R}^{d_v*d}$ of the linear projection. We will keep the same notations as \cite{brunner:etal:2020}, then for ease of understanding, we omit layer and head indices in the upcoming proofs since they remain valid for each head and layer in transformer models.
\begin{equation}
Attention(Q,K,V)H =  (A.V).H = 
\end{equation}
$$A.(E.W^{V}).H = A.T$$
where the attention matrix A is defined as: 
$$ A = softmax(\frac{QK'}{\sqrt{d_q}}) \in \mathbb{R}^{d_s*d_s}$$ and $Q \in \mathbb{R}^{d_s*d_q}$ is the query matrix, $K \in \mathbb{R}^{d_s*d_q}$ is the key matrix, $V=E.W^{V} \in \mathbb{R}^{d_s*d_v}$ is the value matrix (the bias is assumed to be zero), $ W^{V}\in \mathbb{R}^{d*d_v}$ is the matrix that projects embeddings $E$ to the value matrix V, $H \in \mathbb{R}^{d_v*d}$ is the matrix corresponding to the linear layer  that reshapes the concatenation of contextualization matrices into the same dimension as our inputs vectors and $T= V.H$ is a matrix considered to simplify notations, with $d_s$ corresponding to the input sequence length, $d$ to embedding dimension and $d_v$ to the attention head dimension. 

Concerning the identifiability of AW, consider the function $f: A\mapsto AT$. We note that $f$ is linear with respect to A, then f is injective $\Leftrightarrow Ker(f) = \{0\}$. Let `` $'$ `` stand for the transpose of a matrix.  We now study the injectivity of $f$ : for $A \in Ker(f),  f(A) = 0 \Leftrightarrow AT = 0 \Leftrightarrow T'A' = 0$. 
The following proposition establishes necessary and sufficient conditions for injectivity:
\begin{prop} \label{injective}
$f$ is injective iff $Ker(T') = \{0_{d_s}\}$
\end{prop}
\noindent
For a full proof of Proposition \ref{injective}, see Appendix A.
\hidden{
%Let's now see when injectivity fails in transformers.  
\cite{brunner:etal:2020} provide a sufficient condition for the non-identifiability or loss of injectivity of $f_A$.
\begin{prop}  \label{sufficient}
A sufficient condition for $f_A$ to fail to be 1-1 and for AW to be nonidentifiable is that: $d_s > d_v$  
\end{prop} 
\noindent
A full proof of Proposition \ref{sufficient} is in Appendix 2.
}
\subsection{Effective Attention restores identifiability}
\cite{brunner:etal:2020} shows that a sufficient condition for $f$ to fail to be 1-1 and for AW to be nonidentifiable is that: $d_s > d_v$. 
To restore identifiability, \cite{brunner:etal:2020}, 
%based on their result in Proposition \ref{sufficient}
 proposed a solution, effective attention, which consists of the decomposition of each row of AW $A$ into the component in Ker(T') and the component orthogonal to the null space. As all possible AW and effective AW are in finite dimensions $\mathbb{R}^{n}$, then we can write: $ \mathbb{R}^{d_s} = Ker(T)'^\bot + Ker(T')$. Then $AT = (proj_{Ker(T')} (A) + proj_{Ker(T')^{\perp}} (A))T$ $= proj_{Ker(T')^{\perp}} (A)T = (A - proj_{Ker(T')} (A))T $. \\ %$proj_{Ker(T')(A)}$ is calculated by Singular Value Decomposition or singular vectors.\\
There are two problems with the solution proposed by \cite{brunner:etal:2020}.  First, elements of $Ker(T')$ might include negative weights. In addition even if these constraints hold,
there is no guarantee that for $A \in Ker(T')$ : $A \geq 0$ and $A1 =1.$ \cite{brunner:etal:2020}'s solution does not respect probability constraints; and as a result, their effective attention will not lead to easily interpretable elements.  We will tackle these two problems separately. \\

\section{A new solution for Identifiability:Efficient Attention}
In the transformer models typically considered in the literature, $A$ is the result of a softmax, hence its rows are constrained by: $A \geq 0$, and $A1 = 1$, where $1 \in R^{ds}$ is the unit vector of all ones. The condition for nonidentifiability shows that there are infinite number of distributions that verify the same output, but it does not establish that there are several matrices that verify the softmax constraints.  So we need to project attention weights into a different space.  Proposition \ref{suff-nonident} gives details on that space.

%{\color{teal}We now show a sufficient condition for $A$ to be non-identifiable, and to verify softmax constraints.}
\begin{prop} \label{suff-nonident}
$d_s > d_v + 1$ suffices for $A$ to be non identifiable and for the null space of $[T,1]$ to be non-zero.\footnote{$[T,1]$ is the matrix formed by adding a column vector of ones to $T$} 
\end{prop}
\noindent
A full proof of Proposition \ref{suff-nonident} is in Appendix A.

Using Proposition \ref{suff-nonident}, we use instead of \cite{brunner:etal:2020}'s $Ker(T')^{\perp}$, a projection into the null space $Ker([T,1]')^{\perp}$, which induces the same prediction as original attention matrix ($AT = A^\bot T$) and an additional constraint ($A^\bot 1 = 1$) which is important for $A^\bot$ to be a probability distribution.  To solve the problem with \cite{brunner:etal:2020}'s approach for explanations based on attention, we need to prove that the projection into $Ker([T,1]')^{\perp}$ has the global properties of a probability space and that there is an effective means of calculating $Ker([T,1]')^{\perp}$.

\subsection{Global properties of $Ker([T,1]')^{\perp}$}  Our work relies on the decomposition of each row of attention matrix into two supplementary spaces $Ker([T,1]')$ and its orthogonal. Once again for finite dimensions $\mathbb{R}^{d_s}$, $\mathbb{R}^{d_s} = Ker[T,1]'^\bot + Ker[T,1]'$, which means that any vector of $x \in \mathbb{R}^{d_s}$ can be written with a unique decomposition $x = x^\sharp + x^\bot$, where $x^\sharp$ is the projection of $x$ into $Ker([T,1]')$ and $x^\bot$ into its orthogonal.

Note that $AT = [A_1 T, ..., A_{d_s} T]$ with $A_i$ are rows of $A$. We do the same decomposition for each row of attention matrix $A$, then: $A = A^\sharp + A^\bot$.  For any matrix $A$, we have $AT= (A^\sharp + A^\bot)T = A^\sharp T + A^\bot T $ but each row $A^\sharp_i \in Ker[T,1]'$, then $A^\sharp T = 0$, which means that $AT = A^\bot T$. %Every  matrix A can be written in the format $A = projection_{(Ker([T,1]')^{\perp})} (A) + projection_{(Ker([T,1]')}(A)$.Thus, with $Ker([T,1]')^{\perp}$:  $AT = (proj_{Ker([T,1]')} (A) + proj_{Ker([T,1]')^{\perp}} (A))T$. 

We also know that since $A$ is a softmax, then $A.1 = 1$, therefore $ A^\bot 1 + A^\sharp 1 = 1$ but each row $(A^\sharp)_i \in Ker[T,1]'$, then $A^\sharp 1 = 0$, so $A1 = A^\bot 1$.  This means that we can identify each attention matrix responsible for an output, by a unique projection that verifies the same output and has the sum of each row equal to one.

%****The decomposition of any finite matrix A (and so all relevant cases for AI) is unique (supplementarity). The reason which motivates the choice of this decomposition is that $Ker([T,1])^{T}$ is the space which assures us identifiability and maintains the properties of an attention matrix as a probability distribution (Proof in Appendix A.3).  The result is an attention distribution that we can compute, and it is provably a probability distribution and on which AW remain identifiable. 
%****}

\subsection{Positivity of efficient attention weights}

While we have proved that the weights of efficient attention matrices have certain properties of probability distributions (they sum to one)
but we have not shown that every weight is positive $w \geq 0$, which is essential for each row in an AW to be a probability distribution.  \\
Let's prove that $A^\bot = proj_{Ker([T,1]')^{\perp}} (A))$ is a probability distribution, under the following conditions: for each row $i \in \{1,..,n\}$ : $ 0 < A_i = A^{\perp}_i + A^\sharp_i \leq 1$, $A^{\perp}.1 = 1$, $ A^\sharp_i.1 = 0$ and $<A^{\perp}_i, A^\sharp_i> = 0$.  These conditions are ensured by the projection. So now  we need to prove is that under those conditions, $A^\bot \geq 0$.\\
\hidden{also, if there are negative values of $a^{\perp}$ they are all include in $\{i, a^\sharp_i > 0\}$ because $a^{\perp}_i + a^\sharp_i > 0$ \\
then when $a^\sharp_i > 0$, $a^\sharp_i > 0$ or $a^\sharp_i < 0$ \\
$ i \in \{i, a^\sharp_i > 0\}$ means that $ i \in I_0 = \{j \in \{1,..,n\} : (a^\sharp)_{i,j} > 0$ and $(a^{\perp})_{i,j} \geq 0\}$ or
$i \in I_1 = \{j \in \{1,..,n\} : (a^\sharp)_{i,j} > 0$ and $a^{\perp}_{i,j} < 0\}$ \\
}
We prove by induction that under the conditions mentioned above $\forall n \geq 2$, $A^\bot$ can't have negative weights.\\
 For $n=2$, we suppose that $A^\bot$ has a negative weight, for simplification, let's suppose the negative weight is $a^\bot_{i,1} < 0$ with $i \in \{1,2\}$. \\
 With constraints we have it means that: $ a^\bot_{i,1} + a^\bot_{i,2} = 1$ and $ a^\sharp_{i,1} + a^\sharp_{i,2} = 0$ then  $a^\bot_{i,2} = 1 - a^\bot_{i,1}$ and $ a^\sharp_{i,2} = - a^\sharp_{i,1}$ \\
 We know also that $a^\bot_{i,1} a^\sharp_{i,1} + a^\bot_{i,2} a^\sharp_{i,2} = 0 $ then \\
 $a^\bot_{i,1} a^\sharp_{i,1} + (1 - a^\bot_{i,1}) (- a^\sharp_{i,1} ) = 0 $, so  $a^\sharp_{i,1} ( a^\bot_{i,1}  - (1 - a^\bot_{i,1})) = 0 $ \\
We know that $ a^\sharp_{i,1} + a^\bot_{i,1} > 0$ then $a^\sharp_{i,1} > -a^\bot_{i,1} > 0$ \\
 Which means that $ a^\bot_{i,1}  - (1 - a^\bot_{i,1})) = 0$, then 
 $ a^\bot_{i,1} = \frac{1}{2} > 0$  which is in contradiction with our supposition. The same reasoning remains true by taking $a^\bot_{i,2}$ instead of $a^\bot_{i,1}$. \\
 We suppose that this property is true for $n \geq 2$, and to prove that it remains true for for $n+1$, we have only to get back to the case of n variables by taking: $a_{i,n+1}^\bot  = 1 - \sum_{j=1,..,n} a_{i,j}^\bot$ and $ a_{i,n+1}^\sharp = -\sum_{j=1,..,n} a_{i,j}^\sharp $ and then use the induction assumption. $\Box$\\ 
%Omar, I have removed non- from the subsection title
%A better solution is to decompose $A$ into the projection of $Ker([T,1]')^{\perp}$:  $AT = (proj_{Ker([T,1]')} (A) + proj_{Ker([T,1]')^{\perp}} (A))T$.  
\subsection{Computing the projection $Ker([T,1]')^{\perp}$}

The final challenge is to determine mathematically $Ker([T,1]')^{\perp}$.  We do this now.  We first prove a new simpler form for the projection and then providing a computable and analytic method to find that simpler form mathematically.  We show how to find this mathematically using the image of $[T,1]$. The image of a linear function $f$, $Im(f)$, is the set of values that the function $f$ can take by applying the linear transformation to elements of its domain.  
\begin{definition}
     For a linear function  $f : V \rightarrow W$ that maps from a vector space  $V$  to a vector space  $W$, we define $Im(f)$ or $f(V)$: $Im(f)= \{f(v) $ : $ v \in V\}$.
    
\end{definition}
\begin{prop}$Ker([T,1]')^{\perp} = Im([T,1])$
\end{prop}
$x \in Ker([T,1]') \Leftrightarrow [T,1]'x = 0 $ $ \Leftrightarrow$ $\forall y \in \mathbb{R}^{d+1} $ : $ \langle [T,1]'x,y\rangle = 0 $   $ \Leftrightarrow$ $\forall y \in \mathbb{R}^{d+1} $ : $ \langle x,[T,1]y\rangle = 0.$    $ \Leftrightarrow x \in Im([T,1])^{\perp}  $ $ \Leftrightarrow Ker([T,1]') = Im([T,1])^{\perp} \Leftrightarrow Ker([T,1]')^{\perp} = Im([T,1])^{\perp.\perp}$, since we are in finite dimensions then $ Im([T,1])^{\perp\perp} = Im([T,1]) $. As a result, $Ker([T,1]')^{\perp} = Im([T,1])$. $\Box$ \\

%We first show that  $Im([T,1]) \subset Ker([T,1]')^{\perp}$ : \\
%$x \in Im([T,1]) \Rightarrow \exists y \in R^{d+1} $ : $x = [T,1]y$ and $z \in Ker([T,1]') \Rightarrow [T,1]'z = 0$ \\
%then $\langle x,z\rangle = \langle [T,1]y,z\rangle  = \langle y,[T,1]'z\rangle = 0 \Rightarrow x \in Ker(T')^{\perp} \Rightarrow Im([T,1]) \subset Ker(T')^{\perp}$ \\
%Let's show now that $ Ker([T,1]')^{\perp} \subset Im([T,1]) \Rightarrow Im([T,1])^{\perp} \subset (Ker([T,1]')^{\perp})^{\perp}  $, we are in finite dimension then  $ (Ker([T,1]')^{\perp})^{\perp}  = Ker([T,1]')$ then let's show that $Im([T,1])^{\perp} \subset Ker([T,1]' $: \\ 
 %$x \in Im([T,1])^{\perp} \Rightarrow \forall y \in R^{d+1}  : \langle x,[T,1]y\rangle = 0 \Rightarrow \forall y \in R^{d+1}  : \langle [T,1]'x,y\rangle = 0 \Rightarrow [T,1]'x = 0 \Rightarrow x \in Ker([T,1]') \Rightarrow Im([T,1]) ^{\perp} \subset Ker([T,1]') $
% $\Box$\\
% Finally \fbox{$Ker([T,1]')^{\perp} = Im([T,1])$}\\
%
 We now show an analytical way to find $Im([T,1])$.\\
 Consider the standard basis of $\mathbb{R}^{d+1} : (e_{1}, ..., e_{d+1})$.  Now $Im([T,1]) = vect([T,1]e_{1}, ... [T,1]e_{d+1})$ gives us a generating family that we can transform into a basis using the Gauss Pivot algorithm and then get $proj_{Im([T,1])} = proj_{Ker([T,1]')^{\perp}}$ to finally have our new efficient matrix $A_{\mbox{\scriptsize{efficient}}} = proj_{Ker([T,1]')^{\perp}}(A)$. This procedure is general and holds for all attention matrices; it provides an effective computation of identifiable AW matrices that are identifiable.% we'll have identifiability of attention distribution. \\
 \\

 {\color{magenta}

}

\begin{prop}When A is identifiable, $A_{\mbox{\scriptsize{efficient}}} = A$
\end{prop}
This follows from the fact that when $A$ is identifiable, $Ker([T,1]'^{T})=\{0\}$, and so $A = proj_{Ker([T,1]')} (A) + proj_{Ker([T,1]')^{\perp}} (A) = proj_{Ker([T,1]')^{\perp}} (A)) = A_{\mathit eff}$ \\

To summarize, efficient attention restores identifiability of AW by removing the weight components that do not influence the generation of the contextualization vectors $AT$, and consequently the model predictions. We can thus extract the distribution that is responsible for the generation of contextualization vectors and which is unique. That is, we can have two different distribution of attention that generate the same prediction, but their efficient attention will be the same :  $A_{1}.T = A_{2}.T$ with $A_{1} \neq A_{2}$ but $proj_{Ker([T,1]')^{\perp}}(A_{1})$ will be equal to $proj_{Ker([T,1]')^{\perp}}(A_{2})$ and thus correspond to $A_{\mathit{eff}}$.  Our technique isolates the factors that can serve an explanatory role, eliminating the noise from AW.

\section{Empirical investigations} \label{sec:empirical}

\begin{table}[h]
\footnotesize{
\begin{tabular}{|l|l|l|l|}
 \hline
 Datasets & \#test & $W(P_{A}, P_{A_{\mbox{\tiny {\em eff}}}})$ & $RS(P_{A},P_{A_{\mbox{\tiny {\em eff}}}})$  \\ [0.5ex] 
 \hline\hline
 IMDB & 4356 & 0.0016 & 0.02  \\ 
 \hline
 AGNEWS & 3798 & 0.0026 & 0.02  \\
 \hline
 SST & 1725 & 0.0039 & 0.03 \\
 \hline
 20NEWS & 357 & 0.0051 & 0.02 \\ [1ex] 
 \hline
\end{tabular}
}
\caption{Comparison between predictions generated by attention matrices $P_A$ and those generated by their efficient attention projections $P_{A_{\mbox{\tiny {\em eff}}}}$. \#test gives the number of test samples; the metrics $W$ and RS are defined below.}
\label{table:1}
\end{table}
In this section, we support our approach with three empirical experiments. Our formal results are directly relevant to the observations of \cite{jain:wallace:2019, wiegreffe:pinter:2019}, who respectively claim that « attention is not explanation » and that « attention is not explanation if you don't need it».  They produce different attention matrices generating the same predictions, and claim shows AW cannot be explanatorily important. We use the same models and datasets as they do and show that efficient attention is identifiable and explanatorily relevant in the cases they consider. % then having two different attention distributions would have generated different outputs, which was not the case. 
To make our case for efficient AW in the strongest terms, we used \cite{wiegreffe:pinter:2019}'s architecture, which is made up of an encoder (LSTM/RNN/MLP), followed by an attention layer, then a decoder(LSTM/RNN/MLP). We note, however, that our experimental set up transfers easily to other transformer architectures.

First, we show that predictions generated by attention matrices and their efficient attention matrices are the same. Second, we show that \cite{wiegreffe:pinter:2019}'s  different attention matrices generating the same predictions have the same efficient attention matrices. Third, we show that two attention matrices with different efficient attention matrices generate different predictions. For practical reasons, the values will be very close rather than the same, given the limits of integer precision and approximation.

Following \cite{jain:wallace:2019, wiegreffe:pinter:2019}, we conducted our experiments on the following datasets: IMDB Large Movie Reviews Corpus \cite{maas2011learning}, in which the task is to predict positive or negative sentiment from movie reviews, AG News Corpus \cite{zhang2015character} to discriminate between world  and business  articles, Stanford Sentiment Treebank (SST) \cite{socher2013recursive} to predict positive or negative sentiment from text, And 20 News Corpus to discriminate between belonging to baseball and hockey stories.  We used \cite{jain:wallace:2019}'s train-test split and dataset versions. Our analysis is done on the test set as for \cite{jain:wallace:2019,wiegreffe:pinter:2019}.  Note that all datasets are in English.

$T$ passes through the attention layer to generate the attention matrix $A$. Then $A.T$ is passed as input to the decoder to generate prediction $P_{A}$. Our manipulations are carried out in the step between the encoder and the decoder, where we calculate efficient attention $A_{\mathit{eff}}$ from $A$ and $T$, then its predictions $P_{A_{\mathit{eff}}}$. Note that even though our results have been demonstrated for transformers, they remain valid for attention-based systems, since the role of attention remains to generate contextualization vectors and that we can extract efficient attention responsible for their values from AW matrix A by projecting it onto the space constructed from hidden states matrix T in the same way.

We used a variety of metrics in our experiments.  %They mostly all concurred on the results we expected: first that predictions of $A$ and predictions of $A_{mathit{eff}}$ are very close; second that $A^{\mathit{adv}}_{mathit{eff}}$ with $A_{mathit{eff}}$ are very close, and third that $A_{mathit{eff}}$ and $A^{\dagger}_{mathit{eff}}$ differ in predictions.  
%The metrics differed from one experiment to the next given that we were examining different aspects of the problem.  Here we give the definitions of each one and explain its use.  
First, "Earth-Mover" distance or Wasserstein-1 : $W(P_{A}, P_{A_{\mbox{\tiny {\em eff}}}})$, is used to compare probability distributions \cite{arjovsky2017wasserstein,panaretos:zemel:2019}.  
 Since predictions as well as both original attention $A$ and its efficient projection $A_{\mbox{\em eff}}$ are distributions, this is a natural measure.  %It comes from optimal transport theory \cite{peyre2019computational,villani:2021} that provides a measure for how hard it is to transform one probability distribution into another.   %It is defined as follows: 
$$W(P_{A}, P_{A_{\mbox{\tiny {\em eff}}}}) =  \inf_{\gamma \in \Pi(P_{A}, P_{A_{\mbox{\tiny {\em eff}}}})} \mathbb{E}_{(x,y) \sim \gamma} [||x-y||]$$ 
where $\Pi(P_{A}, P_{A_{\mbox{\tiny eff}}})$ is the set of all joint distributions $\gamma(x,y)$ whose marginals are respectively $P_{A}$ and $P_{{A_{\mbox{\tiny {\em eff}}}}}$. 
Intuitively, $\gamma(P_{A}, P_{A_{\mbox{\tiny {\em eff}}}})$ indicates how much “mass” must be transported from $x$ to $y$ in order to transform the distributions $P_{A}$ into the distribution $P_{A_{\mbox{\tiny {\em eff}}}}$. %The EM distance then is the “cost” of the optimal transport plan.
% Def prise de Scipy mais les deux sont equivalentes
%Define2: that $ W_{(u,v)} = inf_{\pi \in \Gamma_{(u,v)}} \int_{\mathbb{R}^{2}}|x-y| \,d\pi_{(x,y)}$ where $\Gamma_{(u,v)}$ is the set of probability distributions on $\mathbb{R}$x$\mathbb{R}$ whose marginals are $u$ and $v$ on the first and second factors respectively. If $U$ and $V$ are the respective cumulative distribution function of U and V, this distance also equals to:  $W_{(u,v)} = \int_{\mathbb{R}} |U-V|$ 
%The input distributions can be empirical, therefore coming from samples whose values are effectively inputs of the function, or they can be seen as generalized functions, in which case they are weighted sums of Dirac delta functions located at the specified values.  
We also use a Mean Wasserstein distance over the set of matrices for a given database in Tables 2 and \ref{table:3}, as well as the tables in the appendix.

A second measure we use is root-mean-square deviation, $RS(P_{A},P_{A_{\mbox{\tiny {\em eff}}}})$, a measure of accuracy, to compare forecasting errors of different models on a particular dataset.
$$ RS(P_{A},P_{A_{\mbox{\tiny {\em eff}}}}) = \sqrt{\frac{ \sum_{i=1}^{n}
(P_{A_{i}} - P_{A_{i_{\mbox{\tiny {\em eff}}}}})^{2}}{n}}$$  We also define and use Pierson ($r^{2}$) and weighted $L_2$ distance in the Appendix B.

% $r2_{A^{adv}_{\mbox{\em eff}}}(P_{A_{\mbox{\em eff}}}, P_{A_{\mbox{\em eff}}}^{adv})$

% $rmse^{adv}_{\mbox{\em eff}}(P_{A_{\mbox{\em eff}}}, P_{A{eff}}^{adv})$

In all the experiments we are not talking about strict identity between outcomes and shifted attention matrices.  Given the approximation errors due to calculations of projection, we will have approximate values.  But the differences should be very small.  If that is the case, as we noted in Section \ref{sec:background}, this will suffice for identifiablity of AW.  Similar remarks apply to the causal efficacy of AW.

%This is indeed the case.  First, we show that different AW for some of the prior experiments have a common efficient attention core, provided we have an accurate computation of $T$ for the model.  In this experiment, $T$ is the sequence of hidden states, and $A$ is from IMDB and movie classification task from Wiegreffe and Pinter.  As a first illustration we show for $A.T$ from $Eff(A)$ and that $\frac{1}{N}\sum^N_{n = 1}\| Eff(A^n).T - A^n.T \|_2 \leq 10^{-4}$. *** we need individual norms too.***

For our first experiment, to see whether AW matrix $A$ and its efficient projection $A_{\mathit eff}$ have the same predictions, we saw that results were very close, indeed identical, up to approximation errors.  To detail our experimental method for each dataset we examined, we extracted for each attention matrix, the efficient attention matrix that corresponds to it.  We
then compared the predictions generated by the efficient attention matrix and the original matrix using three distinct metrics. Table \ref{table:1} shows that the predictions are the same with a few very small variations due to calculation errors, which confirms our assertion that $A \rightarrow P \Rightarrow A_{\mbox{\scriptsize{\em eff}}} \rightarrow P$.  We replicated this finding in Table \ref{table:4} in Appendix B for all the metrics we tried (also Table \ref{table:6} for relative RS comparisons in the Appendix). 

%Second we take two matrices from \cite{wiegreffe:pinter:2019} from the IMDB corpus and show they have a common efficient core.  If $A$ is an original weight, then $A_\dagger$ is their adversarial A weight.  We show on the IMDB corpus that $\frac{1}{N}\sum^N_{n = 1}\| Eff(A^n).T - Eff(A_\dagger^n.T \|_2 \leq 10^{-4}$.  We note that our algorithm is an adversarial algorithm providing a second A matrix with the same result.

For our second experiment, where we compare original and adversarial matrices of \cite{wiegreffe:pinter:2019}, for all datasets, we apply the adversarial algorithm of \cite{wiegreffe:pinter:2019}, so that for each sample we have its attention matrix and its adversarial attention matrix, we extract the efficient matrix of each and compare them. As can be seen from the results, in cases where the adversarial algorithm generates predictions  close to the values of the original predictions, the effective attention weights of the two are close. Note also that the algorithm proposed by  \cite{wiegreffe:pinter:2019} depends on the number of datasets, as can be seen very clearly in the table that the quality of predictions is degraded when the number of datasets is reduced, as in the case of 20News, and as a result there will obviously be a difference between the original efficient matrices and adversarial efficient matrices.  We also note that our algorithm can also be seen as an alternative way of generating adversarial matrices and that is not depending on the length of dataset.  Our results are summarized in Table 2, where once again the differences between the two efficient matrices is negligible. More extensive tests with other metrics can be found in Table \ref{table:5} in the appendix. 

\begin{table}[h]
\small{
\begin{tabular}{|l|l|l|l|}
 \hline
 Datasets  &   WA-eff & W-Pred& RS-Pred   \\ 
 \hline\hline
 IMDB &  0.0047 & 0.0034 & 0.08   \\ 
 \hline
 AGNEWS  & 0.0066 & 0.0021 & 0.05    \\
 \hline
 SST & 0.0288  & 0.0055 & 0.10   \\
 \hline
 20NEWS  & 0.0236 & 0.0509 & 0.20     \\ [1ex] 
 \hline
\end{tabular}
\caption{WA-eff: $W_M(A_{\mbox{\tiny {\em eff}}}, A^{adv}_{\mbox{\tiny {\em eff}}}))$, Mean Wasserstein distances between efficient attention matrices generated by the attention matrix and those generated by their adversarial attention matrices. W-Pred, RS-Pred:comparisons between their predictions $P_{A_{\mbox{\tiny {\em eff}}}}$ and $P_{A_{\mbox{\tiny {\em eff}}}^{adv}}$ using Wasserstein, $W(P_{A_{\mbox{\tiny {\em eff}}}}, P_{A_{\mbox{\tiny {\em eff}}}^{adv}})$, and RS, $RS(P_{A_{\mbox{\tiny {\em eff}}}}, P_{A^{adv}_{\mbox{\tiny {\em eff}}}})$.}}
\label{table:2}
\end{table}

To complete our study of adversarial AW and efficient AW, Table \ref{table:7} in Appendix B gives a full comparison using Wasserstein distances for the predictions raw AW and efficient AW, predictions of raw AW and adversarial AW from \cite{wiegreffe:pinter:2019}, adversarial AW and the efficient projection of those adversarial weights, and finally the predictions of efficient AW and the projection of adversarial AW.
 
Third, we show that shifting efficient attention in a counterfactual intervention on the model causes a difference in prediction.  We note first that efficient attention separates AW matrices more distinctly than the adversarial operation of \cite{wei:etal:2023}. To illustrate this,we generate for each sample its attention matrix $A$, its efficient attention projection $A_{\mbox{\tiny {\em eff}}}$ and prediction $P_{A_{\mbox{\tiny {\em eff}}}}$, simultaneously, we take $1-A$, we generate its efficient attention projection $(1 - A)_{\mbox{\tiny {\em eff}}}$ and prediction $P_{(1- A)_{\mbox{\tiny {\em eff}}}}$. We tested this procedure on all our datasets, and found that each time efficient attention matrices are different, predictions are different, which what we predicted mathematically. Table \ref{table:3} shows the results with Wasserstein distances between predictions of distinct efficient AW is approximately 10x larger for individual AW and the mean Wasserstein difference on distinct efficient matrices is much larger than that for  %mathematically if $(A^1{\mbox{\tiny {\em eff}}) \neg eff(A^2)$, then $Eff(A_1).T \neq Eff(A_2).T$. 
%To illustrate this, we could have taken only one attention matrix, get its projection $A_{\mbox{\tiny {\em eff}}}$, then take another attention matrix, get its projection $A_{\mbox{\tiny {\em eff}}}$, compare the efficient attention matrices and then the prediction. To support our reasoning, we'll take not just one example but the 20News dataset, then do this operation for each sample (Table ~\ref{table:3}). For information, $ W(P_{A_{\mbox{\tiny {\em eff}}}^{1}}, P_{A_{\mbox{\tiny {\em eff}}}^{2}}) = 0.18$ clearly show that they are different, since we are talking about a probability distribution, and note also the value found is far superior to that found for 
$ W(P_{A_{\mbox{\tiny {\em eff}}}}, P_{A_{\mbox{\tiny {\em eff}}}^{adv}}) $ for our datasets.

\begin{table}[!h]
\small{
\begin{tabular}{|l|l|l|}
 \hline
 Datasets & $  mean(W(A^{1}_{\mbox{\tiny {\em eff}}}, A^{2}_{\mbox{\tiny {\em eff}}})) $ & $ W(P_{A_{\mbox{\tiny {\em eff}}}^{1}}, P_{A_{\mbox{\tiny {\em eff}}}^{2}}) $ \\ 
 \hline\hline
 IMDB  & 0.98 & 0.13     \\
 \hline
 AGNEWS & 0.94 &  0.06   \\
 \hline
 SST & 0.88 &  0.16   \\
 \hline
 20NEWS & 0.97 &  0.18  \\ [1ex] 
 \hline
\end{tabular}
}
\caption{Comparison to show that different efficient attention matrices generates different predictions.}
\label{table:3}
\end{table}

Tables \ref{table:6} and \ref{table:7} below give an overview of RS and Wasserstein distances on predictions of various AW.
\begin{table*}
\centering
\begin{tabular}{|l|l|l|l|l|}
 \hline
 Datasets & $RS(P_{A}, P_{A^{adv}})$   & $RS(P_{A}, P_{A_{\mbox{\tiny {\em eff}}}})$ & $RS(P_{A^{adv}}, P_{A^{adv}_{\mbox{\tiny {\em eff}}}})$  & $RS(P_{A_{\mbox{\tiny {\em eff}}}}, P_{A_{\mbox{\tiny {\em eff}}}^{adv}})$ \\ [0.5ex] 
 \hline\hline
 IMDB & 0.03 & 0.02 & 0.07 & 0.08  \\ 
 \hline
 AGNEWS & 0.01 & 0.02 & 0.04 & 0.05  \\
 \hline
 SST & 0.08 & 0.03 & 0.05 & 0.10\\
 \hline
 20NEWS & 0.20 & 0.02 & 0.03 & 0.20  \\ [1ex] 
 \hline
\end{tabular}
\caption{Comparison between RS values.}
\label{table:6}
\end{table*}

\begin{table*}
\centering
\begin{tabular}{|l|l|l|l|l|}
 \hline
 Datasets &  $W(P_{A}, P_{A_{\mbox{\tiny {\em eff}}}})$ & $W(P_{A}, P_{A^{adv}})$   & $W(P_{A^{adv}}, P_{A_{\mbox{\tiny {\em eff}}}^{adv}})$  & $W(P_{A_{\mbox{\tiny {\em eff}}}}, P_{A_{\mbox{\tiny {\em eff}}}^{adv}})$ \\ [0.5ex] 
 \hline\hline
 IMDB & 0.0016 & 0.0038 & 0.0020 & 0.0034  \\ 
 \hline
 AGNEWS & 0.0026 & 0.0020 & 0.0032 & 0.0021  \\
 \hline
 SST & 0.0039 & 0.0058 & 0.0036 & 0.0055\\
 \hline
 20NEWS & 0.0051 & 0.0481 & 0.0067 & 0.0509  \\ [1ex] 
 \hline
\end{tabular}
\caption{Comparison of  Wasserstein distances}
\label{table:7}
\end{table*}
%To illustrate this, we take the IMDB dataset, from which we extract the effective attention matrices, then we generate random probability distributions, extract their efficient attention matrices and check each time that they are different from the original, then we generate the predictions and see in the results that they are different.

%We illustrate this on the IMDB AW in Appendix ****.

\section{Discussion} \label{sec:discussion}
\begin{table*}
\centering
\begin{tabular}{|l|l|l|l|l|l|}
 \hline
 Datasets & $r^{2}(P_{A}, P_{A_{\mbox{\tiny {\em eff}}}})$ & $mse(P_{A}, P_{A_{\mbox{\tiny {\em eff}}}})$  & $L_2^1(P(A),P(A_{\mbox{\tiny {\em eff}}})$ %$m(\frac{\lVert P(A) - P(A_{\mbox{\tiny {\em eff}}})\rVert}{\lVert P(A) \rVert + \lVert P(A_{\mbox{\tiny {\em eff}}}) \rVert} ) $ 
 & $L_2^2(P(A) - P(A_{\mbox{\tiny {\em eff}}})$%m(\frac{\lVert P(A) - P(A_{\mbox{\tiny {\em eff}}})\rVert}{n}$ )  
 \\ [0.5ex] 
 \hline\hline
 IMDB  & 0.99 & 0.0006  & 0.02 & 0.0003 \\ 
 \hline
 AGNEWS  & 0.99 & 0.0006  & 0.01 & 0.0004\\
 \hline
 SST  & 0.98 & 0.0013  & 0.02 & 0.0008 \\
 \hline
 20NEWS  & 0.99 & 0.0005  & 0.01 & 0.0012 \\ [1ex] 
 \hline
\end{tabular}
\caption{Different metrics comparing predictions generated by attention matrices and those generated by their efficient attention projections.}
\label{table:4}
\end{table*}
\label{sec:appendix}
\begin{table*}
\centering
\begin{tabular}{|l|l|l|l|l|l|}
 \hline
 Datasets  &$r^{2}(P_{A}, P_{A}^{adv})$ & $r^{2}(P_{A_{\mbox{\tiny {\em eff}}}}, P_{A^{adv}_{\mbox{\tiny {\em eff}}}})$  & $MSE(P_{A}, P_{A}^{adv})$  & $L_2^1(P(A),P(A^{adv})$  %m%(\frac{\lVert P(A) - P(A^{adv})\rVert}{\lVert P(A) \rVert + \lVert P(A^{adv})\rVert} ) $ 
 &  $L_2^1(A_{\mbox{\tiny {\em eff}}}, A_{\mbox{\tiny {\em eff}}}^{adv})$ %$m(\frac{\lVert A_{\mbox{\tiny {\em eff}}} - A_{\mbox{\tiny {\em eff}}}^{adv}\rVert}{\lVert A_{\mbox{\tiny {\em eff}}} \rVert + \lVert A_{\mbox{\tiny {\em eff}}}^{adv}\rVert} ) $   
 \\ [0.5ex] 
 \hline\hline
 IMDB  & 0.99 & 0.95 & 0.0014 &  0.029 & 0.33  \\ 
 \hline
 AGNEWS  & 0.99 & 0.98 & 0.0001 &  0.009 & 0.15    \\
 \hline
 SST  & 0.94 & 0.91 &  0.0069 &  0.067 & 0.29   \\
 \hline
 20NEWS  & 0.64 & 0.62 & 0.0413 & 0.167 & 0.70 \\ [1ex] 
 \hline
\end{tabular}
\caption{Different metrics comparing attention matrices, its efficient attention matrices, adversarial attention matrices, their efficient attention matrices and their predictions.}
\label{table:5}
\end{table*}
%{\color {magenta} This is where we have to show that the Omar fix disarms the criticism.  Exactly what do we want to do?  We want to show that when we eliminate the noise, the replacement of an attention matrix by another has a counterfactual effect.  Maybe not a complete effect because as Omar has pointed out, the attention matrices really affect the hidden states and only indirectly the output}. \\

Given that identifiability of AW as distributions maximizes their explanatory value in the sense that we explained in Section \ref{sec:background}, our efficient AW provide the strongest explanation possible from attention weight distributions.  Combining this with the experimental results in prior literature that AW are sometimes not identifiable shows that using efficient attention is superior to using either raw attention or \cite{brunner:etal:2020}'s solution for explanatory purposes in many datasets.

A question that arises is, how different is a model $M'$ which differs from a model $M$ only insofar as original AW in $A$ have been replaced by efficient AW $A_{\mathit{eff}}$ in $M'$ ($M' = M[A_{\mathit{eff}}/A]$?  Is $M'$ for instance a surrogate model, an approximation of the original model?  

Given our results, $M'$ is not a surrogate model of $M$.   To see this, note that transformer models are finite automata with (internal) states $S$ and $\Lambda$ labeled transitions $p {\overset {\alpha }{\rightarrow }} q$ for $\alpha \in \Lambda$ for $p,q \in S$. If 2 states are input output equivalent we write $p \equiv q$.
Examples of states could be the output of a given attention layer.  \begin{definition} 
Two finite automata ${\mathcal S} = {\displaystyle S\times \Lambda \times S})$ and ${\mathcal T} = {\displaystyle T \times \Lambda \times T})$ are strongly equivalent iff 
 there is a 1-1 and onto function $\beta: S \rightarrow T$ such that for every $s \in S$ and every $t \in T$, 
(i) if $s\mathrel {\overset {\alpha }{\rightarrow }} s' \in {\cal S}$, then $\exists {\displaystyle t\mathrel {\overset {\alpha }{\rightarrow }} t'} \in {\cal T}$ such that $(s' \equiv t')$; 
(ii) if $ t\mathrel {\overset {\alpha }{\rightarrow } t'} \in {\cal T}$, then $\exists {\displaystyle s \mathrel {\overset {\alpha }{\rightarrow }} s'} \in {\cal S}$ such that $(t' \equiv s')$.
\end{definition}
\noindent
Since $A_{\mathit{eff}}$ and $A$ have the same predictions in the sense that $A_{\mathit{    eff}}.T = A.T$, we see that:
\begin{prop}
Let $M' = M[A_{\mathit{eff}}/A]$ be defined as above.  Then $M$ and $M'$ are strongly equivalent.\\
\end{prop}

%but isolate the causally effective component of the complex model.    We are not approximating the more complex model A but {\bf decomposing} its AW into its effective part (E) and noise.  The model with E is provably {\bf trace or bisimulation equivalent} to A, and so strongly equivalent.  Our work relates to mechanistic interpretability that seeks to explain behaviors of ML models in terms of their internal components.
%***

 Any compelling argument against the usefulness of AW as explanatory devices should exploit our efficient AW.  To this end, we now reconsider the criticisms of AW as explanatory vehicles for NLP tasks involving text classification.  We look at two: the use of uniform weights, the use of adversarial weights using the adversarial technique of \cite{wiegreffe:pinter:2019}. 
 
 Uniformity: Note that the term {\em uniform} here does not mean that all weights have the same value, but that non-zero weights are the same.  \cite{wiegreffe:pinter:2019,vashishth:etal:2019,tutek:snajder:2020} have used uniform weights on contextual elements instead of attention matrix weights to test if attention matrices are important for the output.  They found that uniform weights did not substantially alter the predictions at least for some tasks.  This argument is far from conclusive because it does not look at the nature of the effective components in AW.  % for three reasons.  First,by using uniform weights, they provide a particular score function that associates uniform values. This doesn't in and of itself make attention less important for output; it just changes how they might operate.  
 \cite{brunner:etal:2020}'s effective AW are more uniform, in general,% and less concentrated along the diagonal elements 
 than raw attention, as are our efficient AW, since they are included in the \cite{brunner:etal:2020} effective attention space.  Thus uniform weights at least for classification task matrices do not differ much from those of efficient attention.  We have verified this experimentally; for uniform $A$,
$A_{\mbox{\em eff}}$ also has rather uniform values in our results; our matrices in the tables are uniform.%, d'ailleurs les tableaux c'est pour A uniforme : Ca concorde avec brunner et c'est normal vu que notre espace est un sous espace .

Finally, the uniform values must be close to original matrix values of \cite{wiegreffe:pinter:2019}.  \cite{brunner:etal:2020} show experimentally that the values that are accentuated in raw attention remain those with the greatest value in effective attention but their value decreases to the point of changing only slightly from the other scores. The same holds for our efficient AW rendering the attention weight identifiable makes it close to uniform.  Our experiment 1 (Table \ref{table:1}) also demonstrates that $A$ and $A_{\mathit{eff}}$ have close values. %Nevertheless the ordinal ranking of elements in both effective attention matrices  is preserved in the uniform matrices of \cite{wiegreffe:pinter:2019}.  
This may explain the good performance achieved by the uniform weights set by \cite{wiegreffe:pinter:2019} in some datasets.

%While \cite{wiegreffe:pinter:2019} and others have shown that standard AW may not be identifiable, this doesn't disprove the importance of AW contribution for explaining the model output and for the decision making process. A better test for the explanatory efficacy of AW would be to show that shifting unique AW for inputs makes no difference to the prediction.  
%We have done this.
%{\color {magenta}-> On a montré avec les experiences que A et {A_\mbox{\em eff}} donnent les memes predictions (Table1)}

Adversarial AW. During our experiments, we noticed that the quality of the predictions generated by the adversarial algorithm proposed by \cite{wiegreffe:pinter:2019} depends on the number of samples in each dataset, as can be seen in our metrics in (Table 2) and our appendix tables, especially for 20News and slightly for SST. On the other hand, our efficient attention matrix can also be seen as an adversarial algorithm that generates a different attention matrix with the same prediction as the original, but in this case is independent of the number of samples in each dataset. 

Permutations of AW. \cite{jain:wallace:2019} argue against attention as an explanatory vehicle by showing that permuting the AW only slightly varies the output. However, their attention is a vector not a matrix.  Where AW are true matrices, our experiments have shown the values are relatively uniform, so permutation is not expected to produce large differences in results. % Our response is that if we suppose that efficient AW are generally uniform we know that $A = A_{\mbox{\scriptsize{efficient}}} + A_{ker(T')}$.  Thus, $AT = (T^{T}A^{T})^{T} = [T^{T}a_{1},..., T^{T}a_{ds}]^{T} = [T^{T}(a_{\mbox{\scriptsize{efficient}}_1}+ a_{Ker(T')_1} ),..., T^{T}(a_{\mbox{\scriptsize{efficient}}_{ds}}+ a_{Ker(T')_ds} )]^{T} = [T^{T}a_{\mbox{\scriptsize{efficient}}_1},..., T^{T}a_{\mbox{\scriptsize{efficient}}_{ds}}]^{T}$
%with $a_{1},..,a_{d_s}$ columns of $A^{T}$. Since efficient AW are generally uniform and the values of each element $T^{T}a_{\mbox{\scriptsize{efficient}}_i} $ should be pretty close to each other, then randomly permuting the AW will only change the output a little, which conforms to the result found by \cite{jain:wallace:2019}.

%{\color{magenta} je pense on retire cet argument, parce que la matrice d'attention chez Jain et wallace est un vecteur pas une matrice}

%We have only discussed the results for attention matrices in so called {\em single sequence tasks} like text classification.  \cite{vashishth:etal:2019} have experiments that show substantial drops in performance when it comes to replacing normal AW with uniform ones or permutations in so called {\em multiple sequence tasks} like question answering or sequence to sequence tasks like machine translation.  Since they did not provide code for their experiments, we were unable to verify the characteristics of their attention matrices in those tasks or the effect of our solution for identifiability on those matrices. \\

Finally we offer a detailed comparison with \cite{brunner:etal:2020} that inspired our work.  First, \cite{brunner:etal:2020} investigate identifiability but they don't say why it is important.  We say why identifiability is important for explainability contra \cite{wiegreffe:pinter:2019}. This is crucial for our paper as we are looking at the explanatory potential of AW. \cite{brown:etal:2020} is not explicitly motivated by explanatory concerns but rather by mechanistic interpretability.

 Second, \cite{brunner:etal:2020} propose the orthogonal of Ker(T') to remove weight components that do not influence model predictions; we show that the orthogonal of Ker([T,1]') preserves identifiability and we show both mathematically and empirically that it removes components that do not affect model predictions.

The projection by \cite{brunner:etal:2020} is not guaranteed to be a probability space and they provide examples where it could not be; we show that our projection guarantees that the image of an AW under a projection into the orthogonal of Ker([T,1]') is also an AW. \cite{brunner:etal:2020}'s effective AW are not real AW, according to definitions in the literature on transformer architectures. We also provide and use an algorithmic method for calculating the orthogonal of Ker([T,1]'). 

We verify experimentally that our AW are distinct when results are distinct and that there is 1 efficient AW for distinct raw AW replicating \cite{wiegreffe:pinter:2019}'s empirical set up; \cite{brunner:etal:2020} does not. We compare efficient AW using the well known Wasserstein metric appropriate for probability distributions, \cite{brunner:etal:2020} cannot. This experimental component provides a method for also using efficient AW in faithful and plausible explanations of model behavior. We argue \cite{brunner:etal:2020} cannot do this, since their effective AW are not real AW but unknown objects.

%{\color{magenta} \cite{brunner:etal:2020}'s identifiability rate experiment that they perform for their method is not pertinent for our method since the predictions generated by \cite{brunner:etal:2020}'s effective attention and our efficient attention are the same---we demonstrate this theoretically and also via our experiments.*** On retire plutot ce point ?}

%    \item Our approach provides an explanatory device, being at the same time identifiable and interpretable.

%It is worthwhile to extend these investigations to multiple-sequence tasks. We have examined the results for single-sequence tasks only like text classification. Nevertheless, it is essential to emphasize that our result is a mathematical finding expected to be applicable to all systems based on attention that work the same way.

\section{Conclusions}
We have provided formal results, restoring the potential explanatory power of attention.  We have shown how to render AW effectively identifiable, reducing noise in AW, while also ensuring their interpretability with {\em efficient attention}. Substituting efficient attention for original attention provides a strongly equivalent model.   

%We also have provided a conceptual analysis of the role identifiability in explanations; identifiability in effect operates at several levels.  While it is not expected that individual AWs be identifiable in faithful explanations that provide causal connections, coherent sets of AWs must be.  In addition, identifiability of such coherent sets is a natural and desirable property for AW to have if they are to to provide plausible as well as faithful explanations.

Our novel concept of efficient attention removes an important obstacle to making AW a plausible basis for faithful and plausible explanations.  As we pointed out, however, if a task for a transformer model does not require contextualization or much contextualization, it may be that AW are not causally involved as they are not needed to complete the task.  Our results for efficient attention justify placing more sophisticated learning strategies on top of AW \cite{arous:etal:2021,heo:etal:2020} for explaining model behavior in tasks where contextualization is needed.  

We have illustrated our approach in three empirical experiments over four databases. We compared predictions from an AW and  its efficient projection, and we show that the predictions are very close, and for practical purposes identical given integer and approximation precision. Second, we took the adversarial AW provided by \cite{wiegreffe:pinter:2019} $A^{\mathit{adv}}$ and compared their efficient AW with the efficient AW of the original matrix. %of a matrix $A$ and projected $A^{\mathit{adv}}$ into $A^{\mathit{adv}}_{\mathit{eff}}$.  We then compared $A^{\mathit{adv}}_{\mathit{eff}}$ with $A_{\mathit{eff}}$.
Third, we showed that given two different efficient matrices $A_{\mathit{eff}}$ and $A^{\dagger}_{\mathit{eff}}$ that their projections differ.  

Our mathematical results are general and apply not only to the simplified attention models of \cite{wiegreffe:pinter:2019} and \cite{jain:wallace:2019} but to full transformer models both of the encoder and full kind.  We conducted our experiments on \cite{wiegreffe:pinter:2019} architecture, as it offered a way of having two different attention matrices presenting the same predictions (adversarial technique). This enabled us to show that they shared the same efficient attention matrix in addition to our other experiments.  Space precluded us from examining more modern transformer models empirically, but counterparts of our results here should hold.
 
Given that efficient AW are identifiable from predictions and that we can trace in a transformer model links from input tokens to AW, we can see how shifts in AW occur with shifts in input.  This will also enable linking efficient AW to human annotations of ``rationales'' on input data to test plausibility of the explanations in more detail as in \cite{deyoung:etal:2019}.  We feel this is an illuminating path for future research. 

AW is only part of a transformer model. We have shown why this part is especially important in model performance, but other parts, e.g. the Residual stream, Layer Normalization or the MLP, will also play a part in a complete explanation of the model's performance.
%Are we creating a surrogate model when we replace a given attention weight with its efficient core?  (NO) and we have proved that the two must give equivalent results so at the level of input/output equivalence they are the same; at the level of trace equivalence should be the case as well.  So in terms of thinking of the model as an automaton in a modal space we should have a clear answer here.

 \section{Appendices}
%Here we prove Propositions 1 and 2 and further discuss metrics used in our experiments.

%{\bf A.0. Formal interpretability.}
%Consider two theories $S$ and $T$  and a meaning preserving translation $\tau$. from the language of $S$ ${\cal L}_S$ into the language of $T$, ${\cal L}_T$---$\tau: {\cal L}_S \rightarrow {\cal L}_T$.   
%\begin{definition}
%Let $S$ and $T$ be formal theories.  $S$ is {\em  formally interpretable} in $T$ just in case there is a translation $\tau: {\cal L}_S \rightarrow {\cal L}_T$ such that if $S$ proves ($\vdash$) $\phi$ ($S \vdash \phi$, then $T \vdash \tau(\phi)$). 
%\end{definition}

%åA translation from ${\cal L}_S$ to ${\cal L}_T$, where $\models$ is the notion of model satisfaction \cite{chang:keisler:1973}, is such that: in every model $\mathfrak{A}$ of ${\cal L}_S \cup {\cal L}_T$, $\mathfrak{A} \models \phi \leftrightarrow \tau(\phi)$.

{\bf A.  Proof of Proposition \ref{injective}} \\
To prove the right to left direction, suppose that : $Ker(T') = \{0_{d_s}\}$ \\ 
$A \in Ker(f)$, we consider : $A' = [a_1, ..., a_{d_{s}}]$ with $a_i$ is the i column of A' then : $T'A' = [T'a_1, ..., T'a_{d_{s}}] = [0,..,0]$ $ \Rightarrow \forall i \in {1,..,d_s} : T'a_i = 0$ we know that $Ker(T') = \{0_{d_s}\}$ then $  \forall i \in {1,..,d_s} : a_i = 0 \Rightarrow A' = 0 \Rightarrow A = 0 \Rightarrow Ker(f) = \{0_{d_{s}*{d_s}}\} \Rightarrow$ f is injective (1) \\
To prove the left to right version we use the contrapositive,  $  Ker(T') \neq \{0_{d_s}\} \Rightarrow Ker(f) \neq \{0_{d_{s}*{d_s}}\} $ \\ 
Assume that $  Ker(T') \neq \{0_{d_s}\} \Rightarrow \exists a \in R^{ds}\_\{0\} : T'a =  0$ 
we take the matrix $A' = [a,..,a] \neq 0$ : $T'A'= [T'a,...,T'a] = [0,...,0] \Rightarrow A \in Ker(f) \Rightarrow Ker(f) \neq \{0_{d_{s}*{d_s}}\}$  (2)\\
Finally $Ker(f) = \{0_{d_{s}*{d_s}}\} \Leftrightarrow Ker(T') = \{0_{d_s}\}$, then $f$ is injective iff $Ker(T') = \{0_{d_s}\}$ \\
Consider now $Ker(T')$ : We know that $T \in R^{d_{s}*d}$; so $T' \in R^{d*d_{s}}$. Now apply the Rank Theorem : $dim(Ker(T')) + rg(T') = d_s$. Since we know that $rg(T') = rg(T)$ then $dim(Ker(T')) = d_s - rg(T)$. Finally if $f_A$ is injective, $Ker(T') = \{0\}$ which means that $rg(T) = d_s$. \\
\hidden{
{\bf A.2.  Proof of Proposition \ref{sufficient}} \\
$T = V.H = (E.W^{V}).H \Rightarrow rg(T) \leq min( rg(E), rg(W^{V}), rg(H) )$ \\
$\left\{
\begin{array}{lll}
E \in R^{d_{s}*d} \Rightarrow rg(E) \leq min(d_s,d) \\
W^{V} \in R^{d*d_{v}}  \Rightarrow rg(W^{V}) \leq min(d,d_v)\\
H \in R^{d_{v}*d} : rg(H) \leq min(d_v,d)\\
\end{array}
\right .$\\
$\Rightarrow rg(T) \leq min(d_s,d_v,d)  \leq min(d_s,d_v)$\\
Finally $dim(Ker(T')) = d_s - rg(T) \geq d_s - min(d_s,d_v)$ \\
Then if  $d_s > d_v : dim(Ker(T')) > 0$.  By the injectivity of $f$, AW are non identifiable; moreover there are an infinity of matrices $A$ that give the same output $AT$.
}

{\bf A. Proof of Proposition \ref{suff-nonident}} \\
As $[T,1]' \in R^{(d+1)*d_{s}}$, we apply the rank theorem and so: $dim(Ker([T,1]')) + rk([T,1]') = d_s$. \\
We know $rk([T,1]') = rk([T,1])$, then $dim(Ker([T,1]')) = d_s - rk([T,1])$. So A is identifiable if $d_s = rk([T,1])$, but here we search for a sufficient condition for non-identifiability. \\
$rk([T,1]) \leq rk(T) + 1 \leq min(rk(E), rk(W^V), rk(H)) +1 \leq min(d_{s},d,d_{v}) + 1 \leq max(d_{s},d_{v}) + 1 \leq max(d_{s}+1, d_{v}+1)$. Therefore $dim(Ker([T,1]')) \geq d_{s} - max(d_{s}+1, d_{v}+1) \Rightarrow dim(Ker([T,1]')) \geq  max(-1, ds-dv-1)$ And we know that  $dim(Ker([T,1]')) \geq 0$ finally  $dim(Ker([T,1]')) \geq  max(0, ds-dv-1)$. \\
Then for $d_s > d_v + 1$ :  $ ker([T,1]') \neq \{0\} \Rightarrow \exists B \in ker([T,1]') \neq 0$ that verifies $f_{A+B} = f_A $ and $B . [T, 1] = 0$.\\ 

{\bf B. More metrics and tables}\\

We used MSE, mean squared error, and $r^{2}(P_{A},P_{A_{\mbox{\tiny {\em eff}}}})$ or Pierson distance, the coefficient of determination between $P_{A}$ and $P_{A_{\mbox{\tiny {\em eff}}}}$:\\
$1 - \frac{\sum_{i=1}^{n} (P_{A_{i}} - P_{A_{{\mbox{\tiny {\em eff}}}}} )^{2} }{ \sum_{i=1}^{n} (P_{A_{i}} - mean(P_{A_i}) )^{2}}$ 

We also used two mean, normalized $L_2$ distances, one for matrices and one for predictions:\\
% e.g. $L_2-one$: $mean(\frac{\lVert A_{\mbox{\em eff}} - A_{\mbox{\em eff}}^{adv}\rVert}{\lVert A_{\mbox{\em eff}} \rVert + \lVert A_{\mbox{\em eff}}^{adv}\rVert} ) $ 
 $L_2^1(A_1,A_2)\mbox{\em : mean}(\frac{\lVert A_1 - A_2\rVert}{\lVert A_1 \rVert + \lVert A_2\rVert} )$ and
$L_2^2(A_1,A_2)\mbox{\em: mean}(\frac{\lVert A_1A_2\rVert}{n}$ )
 
 The latter metrics were less robust and sensitive to the size of the database, especially with \cite{wiegreffe:pinter:2019}'s adversarial method.

\begin{ack}
We gratefully acknowledge the support of the EU grant TUPLES as well as the SARER and Summ-RE ANR-20-CE23-0017 grants.
\end{ack}

%\newpage

\bibliography{mybibfile}

\end{document}